\documentclass[10pt,twocolumn,letterpaper]{article}

\usepackage{iccv}
\usepackage{times}
\usepackage{epsfig}
\usepackage{graphicx}
\usepackage{amsmath}
\usepackage{amssymb}
\usepackage{subcaption}
\usepackage{algorithm,algpseudocode}

\usepackage[pagebackref=true,breaklinks=true,letterpaper=true,colorlinks,bookmarks=false]{hyperref}

\iccvfinalcopy 

\def\iccvPaperID{5201} 

\ificcvfinal\fi
\begin{document}

\title{A Camera That CNNs: Towards Embedded Neural Networks on \\ Pixel Processor Arrays}

\author{Laurie Bose$^{1}$ \kern 5mm Jianing Chen$^{2}$ \kern 5mm Stephen J. Carey$^{2}$ \kern 5mm  Piotr Dudek$^{2}$ \kern 5mm Walterio Mayol-Cuevas$^{1}$\\
$^{1}$University of Bristol, Bristol, United Kingdom\\
$^{2}$University of Manchester, Manchester, United Kingdom\\
}

\maketitle

\begin{abstract}
We present a convolutional neural network implementation for pixel processor array (PPA) sensors.
PPA hardware consists of a fine-grained array of general-purpose processing elements, each capable of light capture, data storage, program execution, and communication with neighboring elements.
This allows images to be stored and manipulated directly at the point of light capture, rather than having to transfer images to external processing hardware.
Our CNN approach divides this array up into 4x4 blocks of processing elements, essentially trading-off image resolution for increased local memory capacity per 4x4 "pixel".
We implement parallel operations for image addition, subtraction and bit-shifting images in this 4x4 block format.
Using these components we formulate how to perform ternary weight convolutions upon these images, compactly store results of such convolutions, perform max-pooling, and transfer the resulting sub-sampled data to an attached micro-controller.
We train ternary weight filter CNNs for digit recognition and a simple tracking task, and demonstrate inference of these networks upon the SCAMP5 PPA system. This work represents a first step towards embedding neural network processing capability directly onto the focal plane of a \\sensor.
\end{abstract}
\section{Introduction}

\begin{figure}[t]
    \centering
   \includegraphics[width=0.99\linewidth]{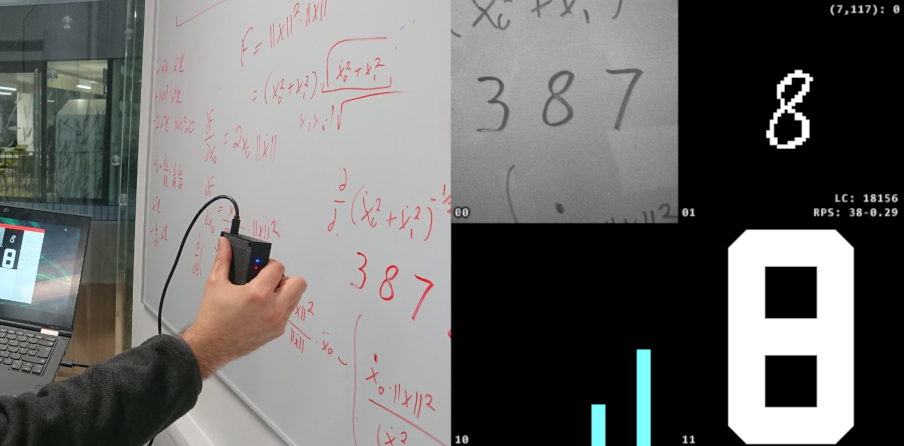}
    \caption{The SCAMP5, a sensor-processor that does true end-to-end capture {\it and} processing with a CNN on its massively parallel pixel-processor array. Here it demonstrates digit prediction from hand-drawn input (left) using inference of a MNIST trained CNN to produce the correct output (bottom-right corner).\label{fig:scamp5_whiteboard}}
\end{figure}  

The application of Convolutional Neural Networks (CNNs) has been done with striding success in a variety of visual tasks.
While most of these applications require significant computational effort, and therefore substantial computer hardware resources (GPUs, FPGAs, cloud-based servers, etc), there is also much interest in applying visual CNN-based inference in scenarios where computing hardware is severely restricted, such as for mobile and small footprint systems. The requirements of neural network algorithms, however, usually surpass the computational capabilities of embedded microprocessors used in these circumstances. This has led to the emergence of a plethora of hardware acceleration engines. The solutions range from mainstream devices adapted for neural network computations, such as GPUs, to custom processor hardware optimised for neural network acceleration \cite{du2015shidiannao, sim2016a142, aimar2018nullhop, efland2016high,chen2016eyeriss,han2016eie}. Bringing the computation closer to the sensor offers distinct advantages in terms of data reduction and power efficiency. This efficiency is essential in many applications, e.g. mobile robots, autonomous vehicles, wearable computing, Internet of Things, among others.

At the same time, a new class of vision sensors is emerging. These devices integrate processors and image sensors in a single integrated circuit. In some cases, the processing circuitry can be incorporated directly into the pixels of the image sensor, resulting in so-called focal-plane processor devices \cite{zarandy2011focal}. Some of these compute relatively simple operations, for example extract temporal contrast in each pixel \cite{posch2011qvga, brandli2014240}. Some implement more elaborate computations, such as convolution kernels \cite{nilchi2009focal}. The advantage of tight sensor-processor integration is the massive bandwidth available at the sensor interface, enabling high rate of operation at low-power, as power-hungry data communications are reduced. At the extreme end of integration of image sensing and processing are Pixel Processor Arrays (PPAs), devices that integrate complete software-programmable processors in each pixel of the image sensor \cite{carey2013100,lopich2013aspa2}. Image computations are carried-out in these processors, and only sparse outputs are transmitted off the sensor device. These devices have been demonstrated to offer unique advantages in applications such as keypoint extraction  \cite{chen2017feature}, depth from focus \cite{ martel2018real}, or visual odometry \cite{bose2017visual}.
In this work, we consider implementations of CNNs on such devices. The flexibility of the compute substrate allows us to contemplate implementing a complete CNN-based classifier in a smart camera system equipped with a PPA device (see Figure \ref{fig:scamp5_whiteboard}). We demonstrate how multiple convolution kernels and max pooling operators can be combined directly on-sensor, to implement neural network computation.

One particular area of interest is that of using low precision weights and neuron activations \cite{venkatesh2017accelerating,muller2015rounding} in order to greatly decrease memory requirements and remove the majority of computational work arising from real value multiplication.
There have been an increasing number of works in this area investigating networks using both binary \cite{courbariaux2015binaryconnect,rastegari2016xnor,lin2015neural} and ternary \cite{li2016ternary,zhu2016trained,alemdar2017ternary} weights, along with implementations of such low precision weight networks on specifically tailored hardware \cite{nurvitadhi2017can,prost2017scalable}.

In this work we propose a novel scheme for ternary weight CNNs on PPA devices and demonstrate inference on a SCAMP5 \cite{chen2018scamp5d} PPA system.
 We propose a suitable network architecture, discuss our training approach, and describe how to implement all the various components required for inference on the PPA itself.
We solve several practical problems related to mapping of computations onto restricted hardware resources of a PPA chip, and demonstrate some simple applications (MNIST digit classification, car tracking). Our experiments are carried out using a SCAMP-5 vision sensor, but the results are applicable to the emerging class of PPA devices in general. The intention of this work is to demonstrate the feasibility of this approach, and pave the way towards future on-sensor vision computations, with even more capable PPA implementations.

\section{Algorithms and Implementation \label{sec:16bit_dreg_images}}

In brief, we take each captured gray-scale image (stored in the PPA's analog registers) and convert it into a specific digital register format, lowering the spatial resolution but retaining a high bit-depth per pixel.
We then perform image convolutions in this format using ternary weight kernels, whose weights correspond to image addition and subtraction operations.
The resulting images from these convolutions are then converted back to analog and stored alongside one another, before undergoing parallel max-pooling.
All the above steps are performed upon the PPA's pixel array, constituting the majority of the computation for inference.
Sparse readout of the array is then used to transfer specific max-pooled data to an attached micro-controller upon which a final fully connected layer is conducted.
The system then outputs the neuron activations of this final layer.


\subsection{Pixel Processor Array}

\begin{figure}[t]
    \centering
  \includegraphics[width=0.99\linewidth]{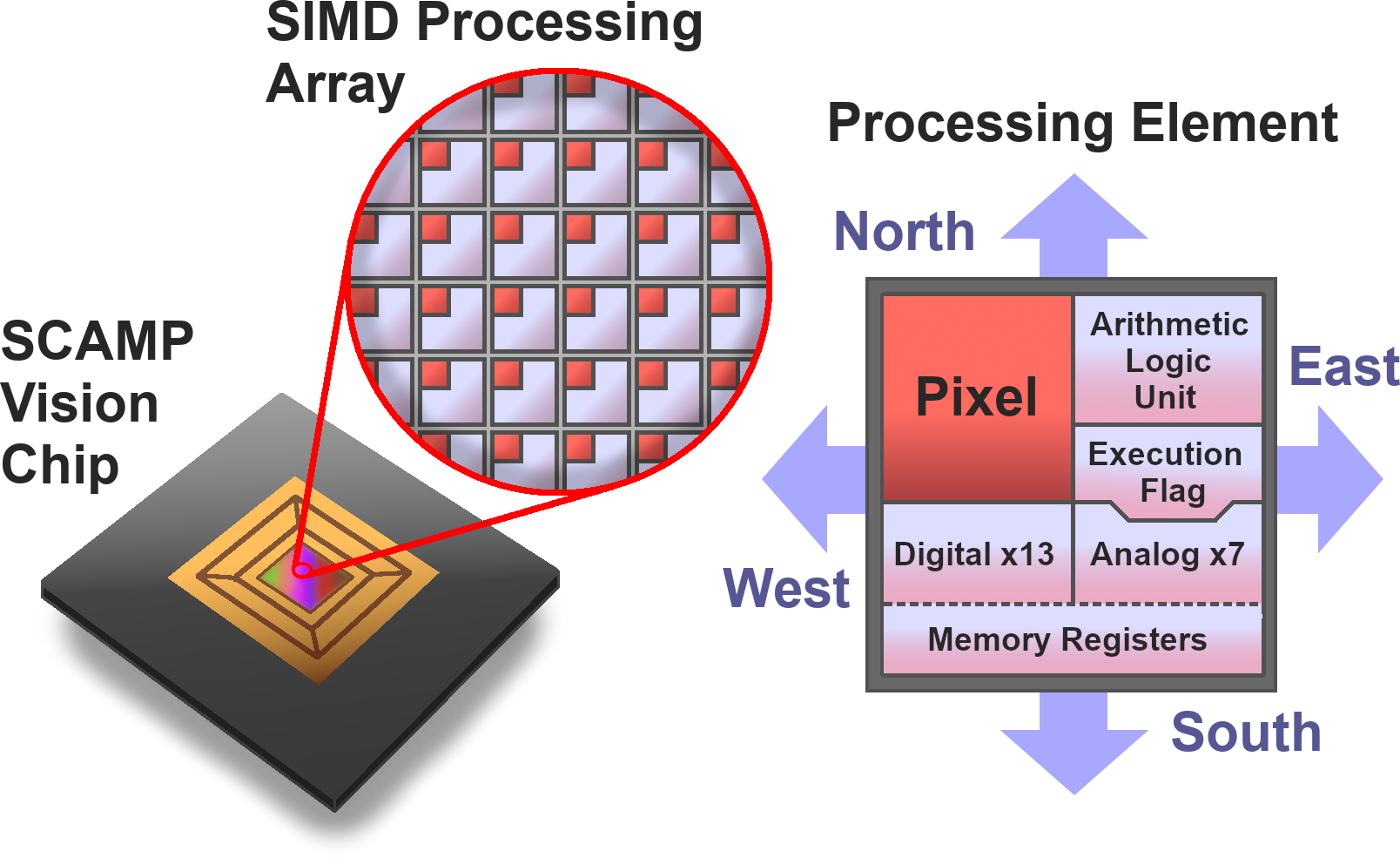}
    \caption{The SCAMP-5 vision chip performs pixel-parallel computations in a SIMD array. Each pixel contains analog and digital storage registers and execution units. An ARM micro-controller controls its operation and performs additional sequential computation.
     \label{fig:scamp5_diagram}}
\end{figure}

The PPA used in our experiments is the SCAMP-5 vision system \cite{chen2018scamp5d}. The architecture is illustrated in Figure \ref{fig:scamp5_diagram}. It is representative of a class of PPA devices, where each pixel of the sensor contains processor circuitry. The resources in each pixel are limited, on the SCAMP-5 device each PE contains 13 digital (1-bit) and 7 analogue memory registers, along with some simple arithmetic, logic and control circuits \cite{carey2013100}. The PPA array is under command of a central controller, and effectively operates as an image-wide SIMD co-processor unit.
This allows operations such as gray-scale analog image addition, and logical OR of binary images, to be carried out in a single instruction cycle across the entire 256x256 image array. 
During typical operation images are acquired through photo-sensors on each pixel, information extracted by parallel processing upon the PPA array, and finally data is transmitted off-chip to the controller. The near-sensor processing approach is very efficient. The SCAMP-5's peak computational performance reaches 655 GOPS, with a maximum power consumption of 1.23W (535
GOPS/W), even though the device is manufactured using two decades old 180nm silicon technology. Very significant gains can be made on future devices in terms of increasing compute power and decreasing power consumption.


In this work, we focus on techniques that require only a small number number of bits per pixel. This is important for PPA implementations. The trade-offs between the amount of local memory and the physical size of individual processors (which limits feasible array sizes) result in small amount of local memory typically available in each pixel (e.g. 13-bits on SCAMP-5, 64-bits in \cite{lopich2013aspa2}, 64-bits in  \cite{shi2014simd}). Furthermore, the limited local memory will need to be shared between algorithms in a more elaborate application. Our approach, coping with an extremely restricted number of bits per pixel, should therefore transfer easily to future digital PPA devices which would allow both both deeper networks and faster computation speeds.


It should be noted that the SCAMP-5 PPA we use in this work contains both analog and digital registers for image storage. 
Analog operation can provide greater speed and efficiency in some cases, however unless carefully addressed, repeated analog operations can both lead to a build up of noise.
Therefore, unlike \cite{debrunner2019auke}, this work conducts image convolutions using digital registers, while analog registers are used for storage and parallel max-pooling.



\subsection{Low Resolution High Bit Depth Digital Images}  

\begin{figure}[t]
    \centering
      \includegraphics[width=0.31\linewidth]{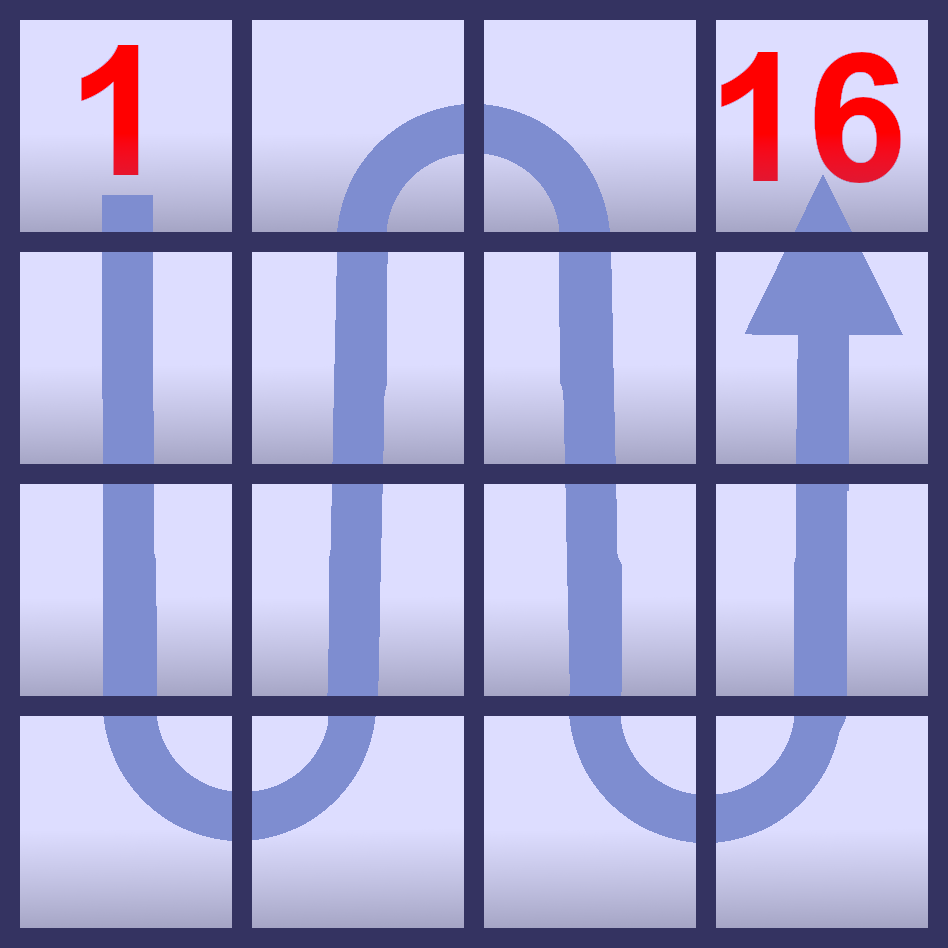}
   \includegraphics[width=0.31\linewidth]{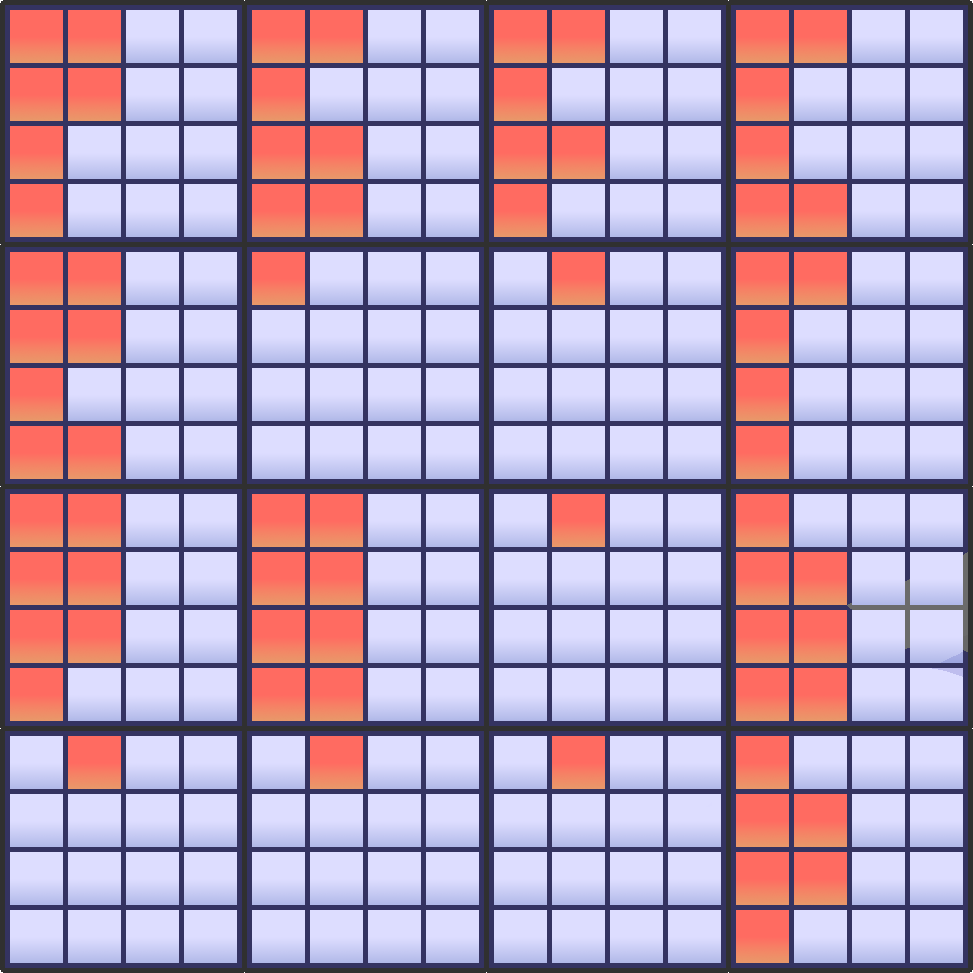}
   \includegraphics[width=0.31\linewidth]{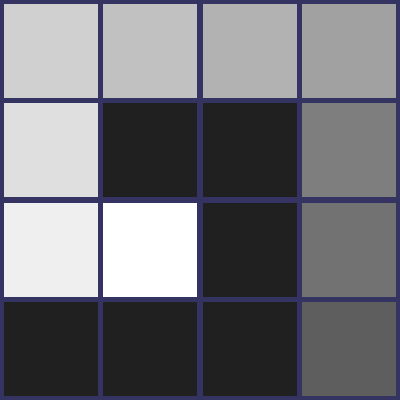}
    \caption{Left the order of bits 1-16 from least to most significant in a 4x4 pixel block. Middle a grid of 16x16 pixels split into 4x4 blocks storing a 16-bit image. Right the same image displayed in gray-scale. \label{fig:4x4_image_format}}
\end{figure}

\begin{figure}[t]
        \centering
        \includegraphics[width=0.95\linewidth]{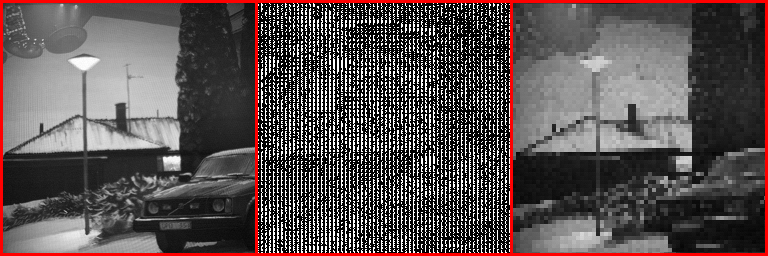}
       \caption{Left to right, an analog image captured by SCAMP5 being converted into the digital 4x4 pixel block format and then converted back into an analog image. Note the decrease in resolution from 256x256 to 64x64.  \label{fig:4x4 block image format example}}
    \end{figure}

    Each of the 256x256 pixel-processors on the SCAMP-5 PPA contains 13 digital (binary) registers. Writing or reading to the same digital register within all pixels of the array thus allows a single 256x256 binary image to be stored and manipulated.
    However, 1-bit images are insufficient for computing and storing image convolution results. 
    On the other hand, using multiple digital registers in unison to store an image of higher bit-depth can tie up a great amount of resources needed for performing other computations.
    To solve this problem, we propose an image format which splits the 256x256 array into 4x4 pixel blocks as illustrated in figure \ref{fig:4x4_image_format} and demonstrated in figure \ref{fig:4x4 block image format example}.
    The 16 digital registers from each 4x4 block (ie each "pixel") are then used to hold a single 16-bit value.
    This digital image format effectively reduces the image storage resolution from 256x256 to 64x64, but increases the bit depth from 1 to 16.
    This provides a better suited trade-off between resolution and bit depth for deep learning tasks, and is used in performing image convolutions.
    Methods of how to efficiently add, subtract and manipulate images stored in this format now follow.
    
 

    \begin{figure}[t]
        \centering
          \includegraphics[width=0.6\linewidth]{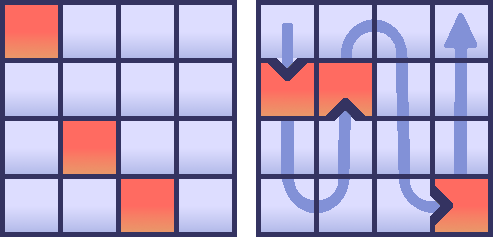}
            \hspace{3mm}
            \includegraphics[width=0.25\linewidth]{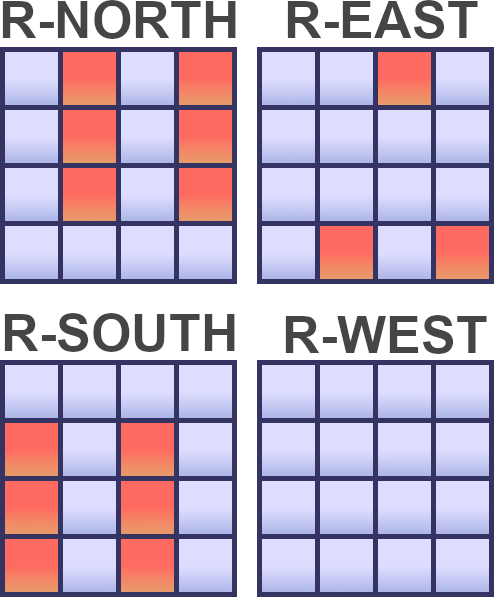}
            
        \vspace{2mm}
        
        \includegraphics[width=0.6\linewidth]{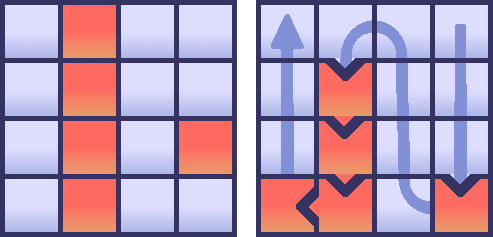}
        \hspace{3mm}
        \includegraphics[width=0.25\linewidth]{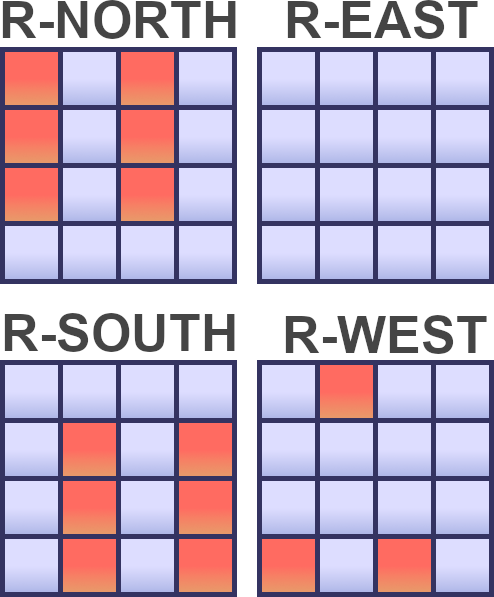}
       \caption{Illustration of 4x4 block bit-shift up (top) and down (bottom) along with corresponding patterns loaded to transfer-direction registers. \label{fig:bit-shift down} \label{fig:bit-shift up}}
    \end{figure}
    
\subsection{Bit Arrangement And Shifting}
    
    The order of bits in a single 4x4 block, from most significant to least significant are shown in figure \ref{fig:4x4_image_format}.
    This arrangement consists of a continuous zigzag path or "bit snake", with each bit location neighbouring both the next and previous most significant locations.
    In SCAMP-5 pixels may only communicate with their four immediate neighbours, and transferring data from one pixel to another located far across the array involves many operations shuffling data from pixel to pixel.
   The proposed bit snake pattern avoids this issue when bit-shifting the image, as each pixel may immediately transfer its data to the location of the next or previous bit. 
   The bit-shifting operations are illustrated in figure \ref{fig:bit-shift up}. 
   The direction in which each pixel transfers its data is determined by four control registers, R-NORTH,R-SOUTH,R-EAST,R-WEST, each specifying a different transfer direction.
    By loading the correct patterns into these registers for each 4x4 "pixel" block, a single data transfer operation can be used to bit-shift the entire image, moving all pixel data forwards or backwards along the bit snake within each 4x4 block.
    This gives us an efficient parallel way to bit-shift images in the 4x4 block format, a vital component for image addition and subtraction.
    
  

\subsection{Addition and Subtraction \label{subsec:add_and_sub}}
Performing image convolutions using ternary weights requires being able to perform a sequence of image additions and subtractions.
This section describes how to perform these operations upon images in the 4x4 block format. 

\subsubsection{Addition}
\begin{figure}[t]
    \centering
   \includegraphics[width=0.9\linewidth]{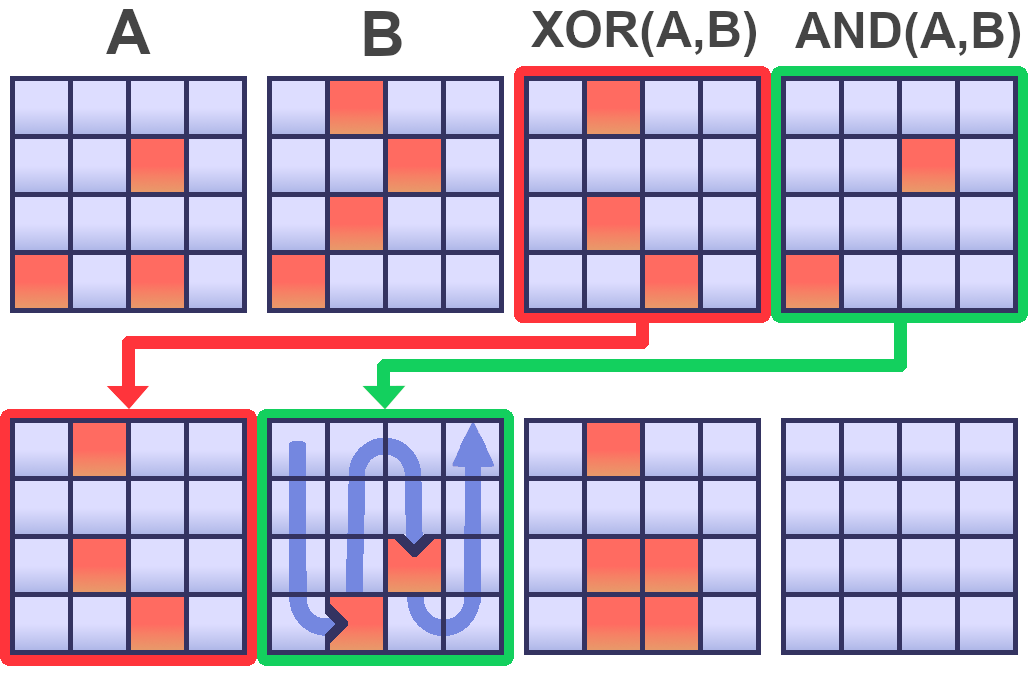}
   \caption{A single 4x4 block addition step. \label{fig:4x4_block_addition}}
\end{figure}

Performing image addition between images A and B involves calculating the two intermediate images AND(A,B) and XOR(A,B).
These can be generated using a combination of the NOR and NOT operation native to SCAMP5 digital registers.
If the content of AND(A,B) is a black (all 0s) image, there are no bits set in the same locations anywhere across A and B, and the result of the A+B is simply XOR(A,B).
However if the image AND(A,B) has set bits within it, then it is copied and bit-shifted up, as the bits set in the same locations across A and B are added together.
Image A is then replaced by A=XOR(A,B), B is replaced by B = BitShiftUp(AND(A,B)), and the entire process is then repeated until AND(A,B) contains no set bits.
This process is illustrated on two 4x4 blocks in figure \ref{fig:4x4_block_addition}.


\subsubsection{Subtraction}

\begin{figure}[t]
    \centering
   \includegraphics[width=0.9\linewidth]{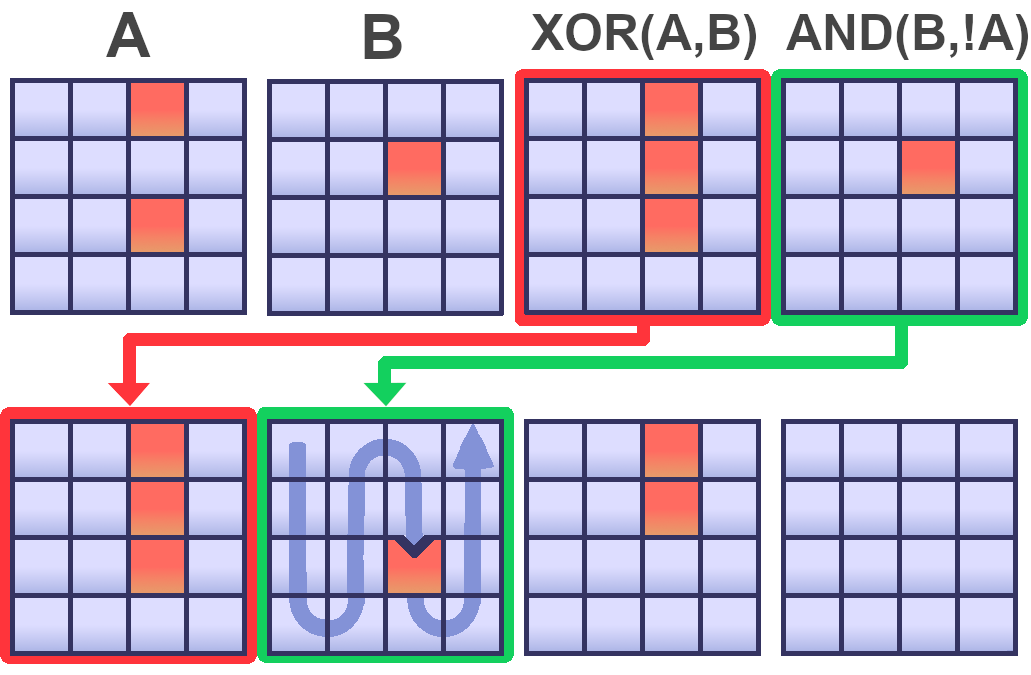}
   \caption{A single 4x4 block subtraction step. \label{fig:4x4_block_sub}}
\end{figure}

The process of subtracting an image B from image A follows a similar set of steps to addition.
The images XOR(A,B) and AND(!A,B) are generated using native NOR and NOT operations, AND(!A,B) then holds the carry for the subtraction, and XOR(A,B) the intermediate result. 
If the carry image has set bits it is bit-shifted up and with image B replaced by B = BitShiftUp(AND(!A,B)) and image A replaced by A = XOR(A,B).
This sequence of steps is then repeated until the carry register is empty of set bits, upon which XOR(A,B) is returned as the subtraction result. 
This process is illustrated in figure \ref{fig:4x4_block_sub}.


\subsection{Checker-Board Storage To Analog Images \label{subsec:areg checker board storage}}
As mentioned previously we avoid using analog registers for performing long computations, however they are used for intermediate image storage of convolution results, since they offer additional local storage resources available in each pixel.
This involves a digital to analog conversion, taking each 4x4 pixel block, and loading an approximation of the block's stored value into a single analog pixel.
This decrease from 16 to 1 array pixels used to store a value allows 16 images in the digital 4x4 block format to be stored within a single 256x256 analog image.
This is done using a checker-boarding scheme as shown in figure \ref{fig:convolution_example}, where every $4^{th}$ pixel in x and y belongs to the same image.


\subsection{Image Convolutions \label{sec:image convolutions}}

Image convolutions generate new images in which each pixel is formed from some linear combination of pixels from a source image.
Specifically the value of each pixel is formed by taking a weighted sum of pixels from a local rectangular region in the source image.
Identical weights are used to form each pixel and the corresponding matrix of weights is referred to as the convolution kernel or filter.

In this work we restrict such kernels to ternary weights (of possibles value 1,0,-1).
This allows image convolutions to be implemented in a parallel manner upon SCAMP, using only image additions and subtractions, avoiding computationally costly multiplication and requiring less memory for storage of weights.
All convolutions are performed between images stored in the 4x4 block digital register   format described in Section \ref{sec:16bit_dreg_images}, making extensive use of the routines laid out for image addition, subtraction and bit-shifting.

\subsubsection{Method}

\begin{figure}[t]
    \centering
    \includegraphics[width=0.47\linewidth]{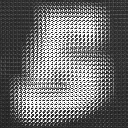}
   \includegraphics[width=0.47\linewidth]{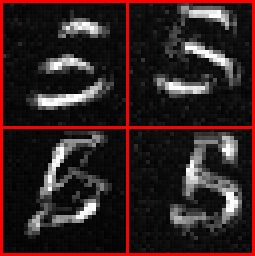}
   \caption{An example of 16 convolution results stored within a single checker-boarded analog register. Four of the convolution images are extracted and shown to the right, \label{fig:convolution_example}}
\end{figure}

Performing a convolution consists of iteratively shifting the source image such that each pixel has visited every location to which it contributes to in the new image.
In our case this involves shifting the entire source image along a zigzag path covering the size of the rectangular kernel being used.
At each step the kernel weight associated with the current shift is examined, and the shifted source image is added to one of two possible images (or ignored on a weight of 0), the first image is used to accumulate the additions of images, and the second to accumulate those images being subtracted.
In the final step the image of accumulated subtractions is subtracted from that of the accumulated additions forming the new image generated by the convolution operation.
This process is outlined in algorithm \ref{alg:convolution} and figure \ref{fig:convolution_example} shows some examples of images resulting from various kernels.

\begin{algorithm}
\begin{algorithmic}[0] 
\State{Clear(B,C)} 
\For{$y = 0 : y < Kernel Size-1$}
    \For{$x = 0 : x < Kernel Size-1$}
        \If{Weight[x][y] == 1}
            \State{B = Add\_4x4\_block\_images(A,B)}
        \ElsIf{Weight[x][y] == -1}
            \State{C = Add\_4x4\_block\_images(A,C)}
        \EndIf
        
        \If{y\%2 == 0}
            \State{Shift\_Image\_East\_x4($A$)}
        \Else
            \State{Shift\_Image\_West\_x4($A$)}
        \EndIf
    \EndFor
    \State{Shift\_Image\_South\_x4($A$)}
\EndFor
\State{A = Subtract\_4x4\_block\_images(B,C)}
\State{Return(A)}
\end{algorithmic}
\protect\caption{$Perform\_Image\_Convolution\_On\_A$ \label{alg:convolution}}
\vspace{-1mm}
\end{algorithm}

%

\begin{figure}[t]
    \centering
     \includegraphics[width=0.95\linewidth]{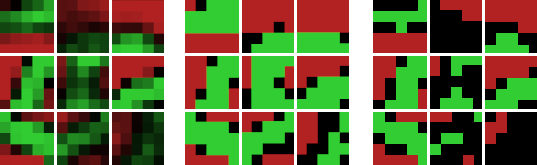}
   \caption{Examples of kernel filters from MNIST training. Left shows the real valued weights, middle and right ternary weight generated with thresholds 0.2 and 0.5 respectively. \label{fig:mnist_filters}}
\end{figure}

\section{Network Training With ternary Weights\label{sec:network_training}}

This section describes the process used to train networks for later inference on the SCAMP5 hardware.
Due to the unique nature of the SCAMP5 hardware which varies significantly from more standard devices we implemented our own software to conduct network training and simulation.
This allowed us to quickly test various ideas and training schemes, accounting for hardware implementation constraints, and to correct any discrepancies between the training on PC and the actual inference on SCAMP5 hardware. 

We restrict ourselves to ternary weights $-1, 0, 1$ on convolutional layers, as these are computed on the pixel array where simple image operations such as addition and subtraction (corresponding to weights $+1,-1$) are greatly preferable to multiplications.
Our approach for training a network involving ternary weights is highly similar to that taken by \cite{courbariaux2015binaryconnect}, extended from binary to ternary weights, and applied to the shared weights of convolutional kernels.
We store real value representations for all weights in the network, during each forward pass step, the real values associated with ternary weights are used to stochastically generate ternary weight values used in that pass.
This discretization is performed according to Equations \ref{eqn:ternary weight probabilty} and \ref{eqn:ternary weight sigma}, which are computationally cheap, hard sigmoid functions.
Essentially the closer a ternary weight's associated real value is to 1, the more likely it will be assigned the ternary value of +1 in the forward pass, and similarly for the values of 0 and -1.
The same ternary weight values generated in the forward pass are then used in the backwards propagation step.
The gradients generated by these ternary values then contribute in updating weights across the network in the parameter update step.
Note we use Rectified Linear Units (ReLU) as the activation function for all neurons and we also bound weights to within the $[-1,1]$ interval, as going outside of these values will cease to affect the process of discretization based upon Equation \ref{eqn:ternary weight sigma}.

\begin{equation}
T(w) =
\begin{cases}
+1 & \text{with probability } \sigma(w) \\
0 & \text{with probability } 1-\sigma(\mid w\mid) \\
-1 & \text{with probability } \sigma(-w)
\end{cases}
\label{eqn:ternary weight probabilty}
\end{equation}

\begin{equation}
\sigma(x) = max(0,min(1,x))
\label{eqn:ternary weight sigma}
\end{equation}

A key insight behind this approach is that with the stochastic discretization process, each ternary weight can be viewed as a noisy approximation of its associated real valued weight.
The gradients generated by these ternary weights, accumulated over many training samples, average out to express the  real values behind them.
This allows stochastic gradient descent to still proceed in a similar manner as on networks with real valued weights.

\section{Inference On SCAMP5\label{sec:inference}}

This section describes the process involved in taking networks trained as per section \ref{sec:network_training} and replicating them upon the SCAMP5 hardware.
As described in section \ref{sec:network_training} each ternary weight has an associated real value, which in each forward pass is used to probabilistically generate an associated ternary value for said weight.
However, when performing inference we need to determine a final fixed ternary value to use for these weights.
This can be done by simply thresholding each ternary weight's associated real value $W^{T}_{R}$ to a ternary value $W^{T}$.
That is for a given threshold  $\alpha \in [0,1]$, a ternary weight $W^{T}$ is assigned a value of $+1$ if $W^{T}_{R} > \alpha$, a value of $-1$ if $W^{T}_{R} < -\alpha$, and $0$ otherwise.
Increasing threshold $\alpha$ results in a greater number of 0 valued weights in the network, decreasing computational cost of inference but potentially at the expense of accuracy as will be explored in later sections.
Figure \ref{fig:mnist_filters} shows an example of ternary weights generated by this thresholding for both high and low values of $\alpha$.

Once all weights have been extracted from the trained network we can execute programs to perform inference of the same CNN architecture directly on the SCAMP5 vision system.
Each program follows a similar scheme of acquiring a camera image, converting it to the digital 4x4 block format described in section \ref{sec:16bit_dreg_images}, and then using the kernel weights stored in FLASH to perform image convolutions upon it as described in section \ref{sec:image convolutions}, storing the resulting images within a checker-boarded analog image as described in \ref{subsec:areg checker board storage}.
Parallel max-pooling can then be performed upon all images stored within this checkerboard image and the resulting low resolution images representing the inputs to the fully connected layer transferred from the SCAMP chip to the ARM micro-controller using a sparse readout of the pixel array.
A final fully connected layer is then performed using these readout images on the ARM core, generating the output layer of neurons.
This final set of neuron activations then makeup the output of the SCAMP vision system, and is transmitted via USB to a PC for visualization.
We now demonstrate such real-time inference in the following section.

\begin{figure*}[t]
    \centering
     \includegraphics[width=0.16\linewidth]{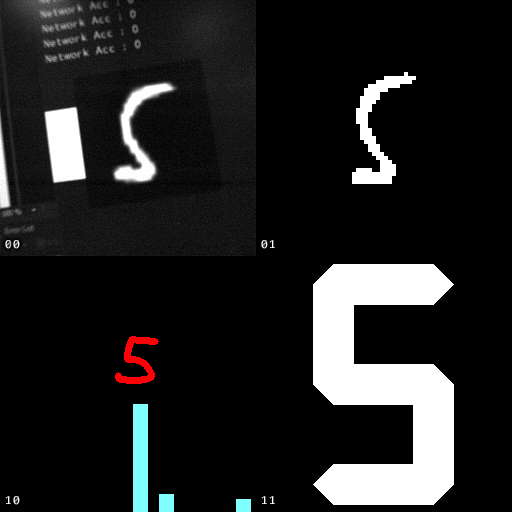}
     \includegraphics[width=0.16\linewidth]{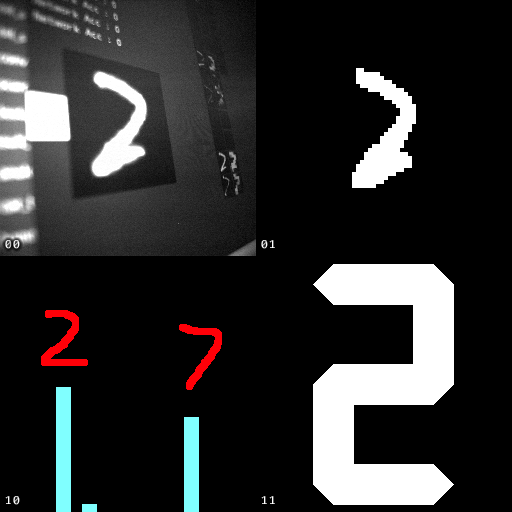}
   \includegraphics[width=0.16\linewidth]{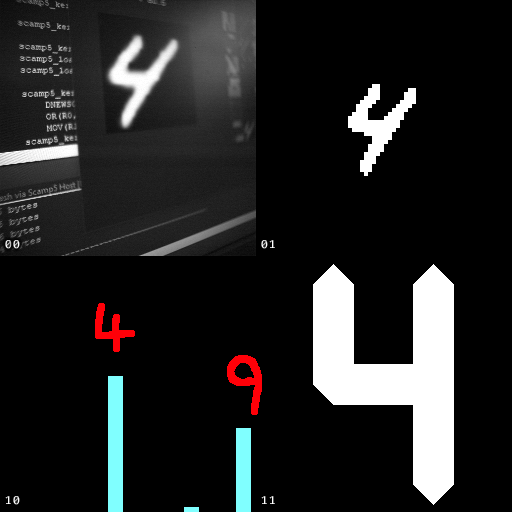}
       \includegraphics[width=0.16\linewidth]{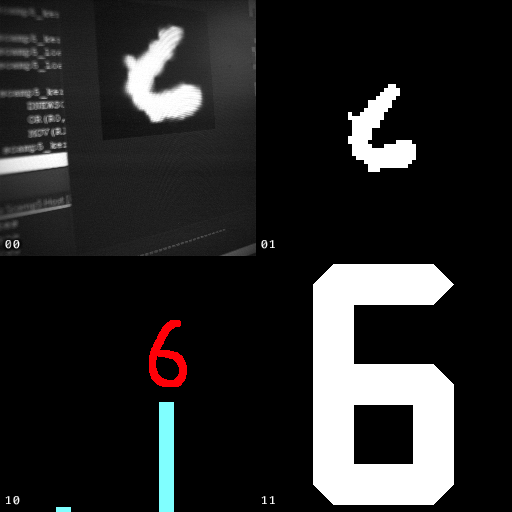}
         \includegraphics[width=0.16\linewidth]{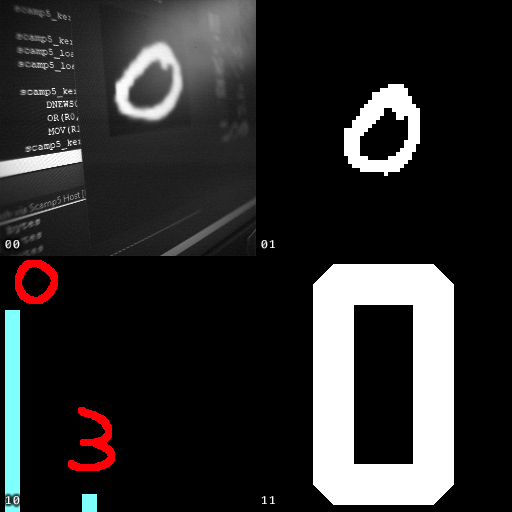}
           \includegraphics[width=0.16\linewidth]{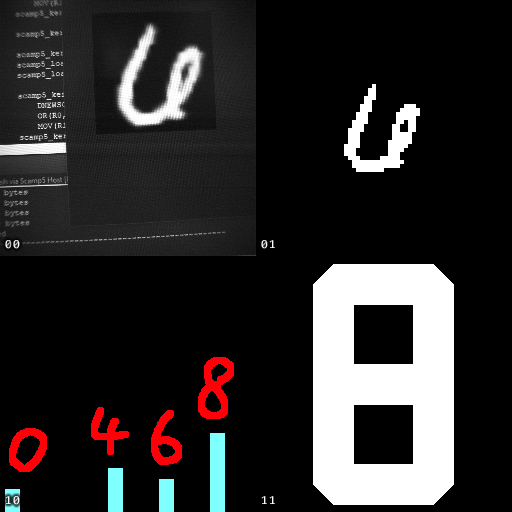}
   \caption{Examples of using the SCAMP5 MNIST trained digit recognition on displayed MNIST digits. Each composite frame contains four panels: top-left is the image captured by the sensor, top-right is the extracted, re-scaled digit image that is fed to the neural network, bottom-left is the classification result i.e. activations of ten output neurons, bottom-right is the abstract visualization of the classification result. Despite the viewing angle distorting the extracted digits the network still achieves correct classification except in highly questionable digits such as that on the right. \label{fig:mnist_digits}}
\end{figure*}

\begin{figure*}[t]
    \centering
          \includegraphics[width=0.16\linewidth]{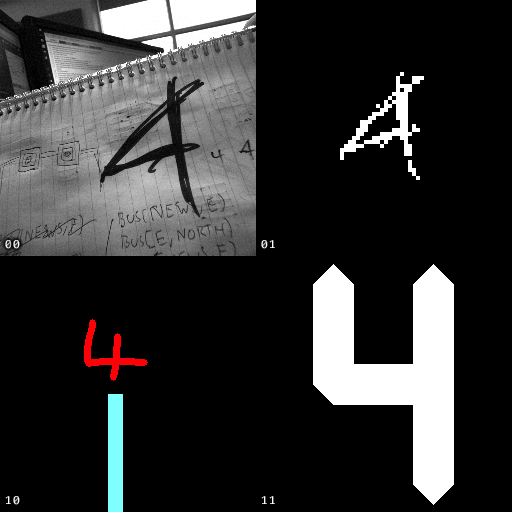}
      \includegraphics[width=0.16\linewidth]{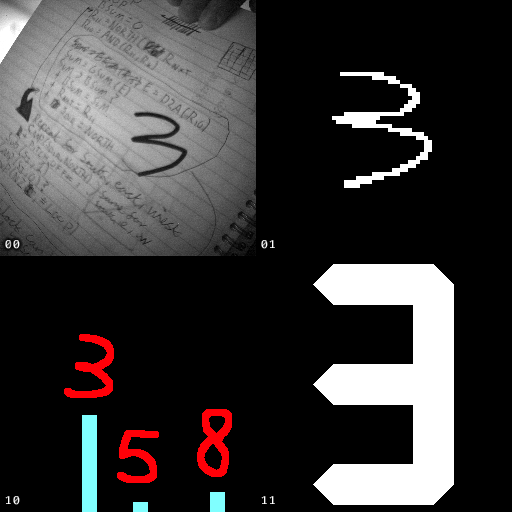}
   \includegraphics[width=0.16\linewidth]{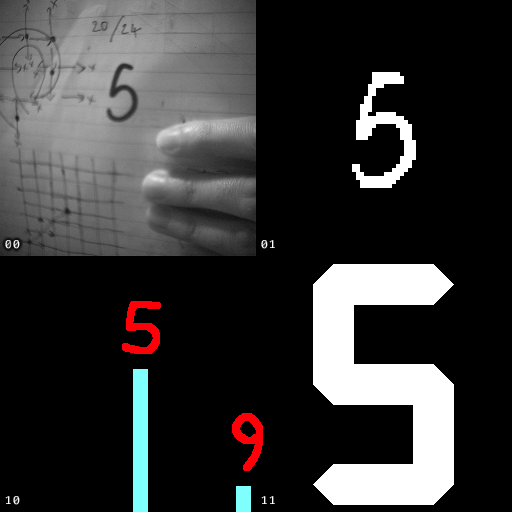}
       \includegraphics[width=0.16\linewidth]{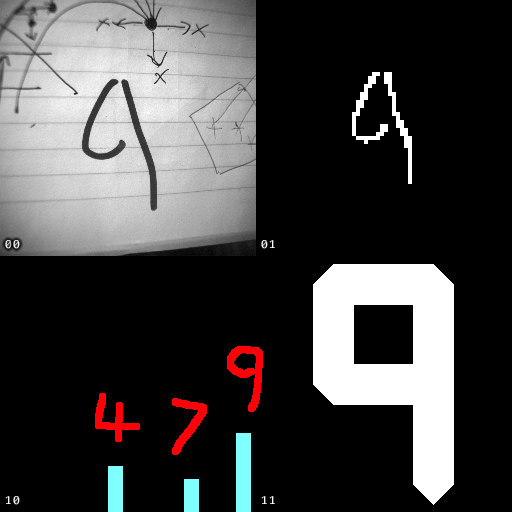}
         \includegraphics[width=0.16\linewidth]{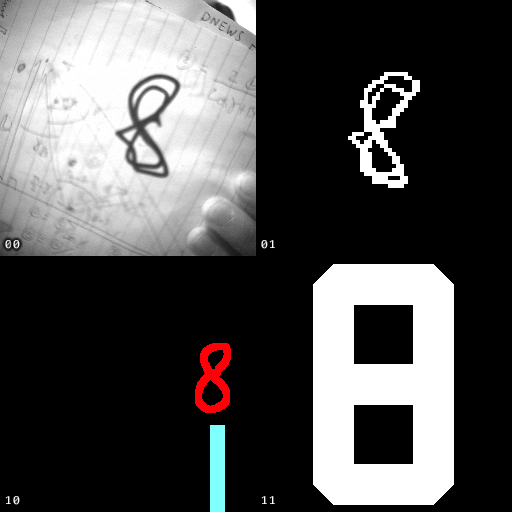}
           \includegraphics[width=0.16\linewidth]{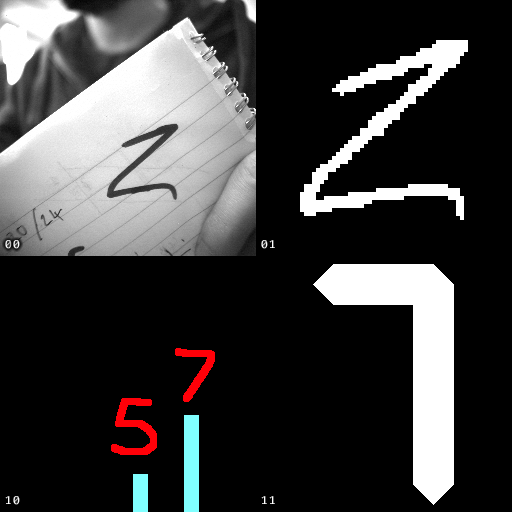}
   \caption{Similar to figure \ref{fig:mnist_digits}, examples of using the SCAMP5 MNIST trained digit recognition on new hand drawn digits. Note the failure case in the right most example where an extremely angular 2 is mis-classified as a 7. The bar plot indicate the activations of the final layer neurons associated with the digits 0-9.\label{fig:custom_digits}}
\end{figure*}

\section {MNIST \label{sec:MNIST}}

\subsection{Data Augmentation and Training}
MNIST images will be displayed on a monitor and captured by the image sensor, as this is the primary input channel for a PPA.
This results in images that deviate from the original MNIST dataset due to viewing angle, sensor position, lighting conditions etc, and so a simple data augmentation was employed to make the network more robust to these variations.
This consisted in applying a random transformation to the image, specifically a translation up to 2 pixels in magnitude, a random rotation between $\pm 20$ degrees, and randomly re-scaling the resulting image by up to $\pm 10\%$ of it's original size separately in x and y.

We trained networks consisting of a single 5x5 kernel convolution layer of 16 filters, followed by a 4x4 max-pooling layer into a fully connected output layer using 8-bit weights. 
Figure \ref{fig:mnist_filters} shows an example of 9 filter kernels resulting from the training process.
Such networks typically obtained a classification accuracy of $\approx95\%$. 

\subsection{SCAMP5 Inference}

When performing an evaluation of MNIST digit recognition inference upon SCAMP5 we face the issue that there is no efficient way to directly input dataset images into the pixel array.
We were instead displaying the test set images to the sensor via a monitor, performing a simple character extraction algorithm on SCAMP5, and feeding the extracted character image through the network and recording the predicted digit to compare against the correct classification.


\begin{figure}[t]
    \centering
   \includegraphics[width=0.9\linewidth]{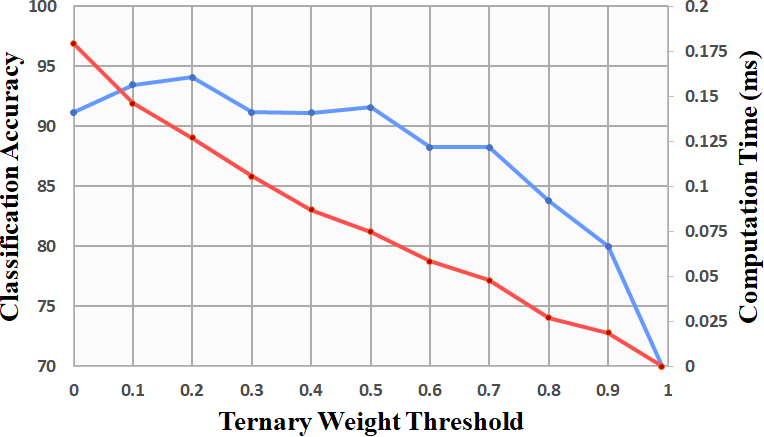}
   \caption{Plot showing both classification accuracy (blue) and convolution computation time (red) against  ternary weight threshold. \label{fig:mnist_results_graph}}
\end{figure}

This character extraction routine locates a white shape enclosed by a black background (or black against white), determines the shape's bounding box, and then extracts the image within this bounding box.
The extracted image is then re-centered and scaled to a desired size, making use of methods described in \cite{bose2017visual} for image transformations on the focal plane.
In the case of MNIST this routine extracted the MNIST character found within the camera image and transformed it to be a similar size and position as characters within the original MNIST images.
A sample of predictions made by the network are shown in Figure \ref{fig:mnist_digits}.

\subsection{Evaluation \label{subsec:MNIST results}}

Figure \ref{fig:mnist_results_graph} shows the classification accuracy of the MNIST network on SCAMP5 and how this accuracy varies across different thresholds used to generate ternary weights from the trained real weight values.
We observed that a threshold of 0, leading to purely binary weights in the network, had lower accuracy and higher image convolution computation time than various non zero thresholds.
This indicates that the extra information allowed by ternary weights is indeed being put to use during classification.
The highest classification accuracy of $94.2\%$ occurred with a threshold value 0.2. 
Note that this is a noticeable drop from the accuracy of $95.4\%$ obtained by the trained network on PC, for which the mismatch between the captured images and training data may partly be responsible.
The computation time to perform a convolution decreased linearly with the threshold used for ternary weights as would be expected since the number of zero valued weights should be approximately proportional to this threshold.
Interestingly with a threshold of $0.99$, where only a couple of weights per convolution kernel are non-zero, the network still had a classification accuracy of $70\%$.
The frame-rate scaled approximately linearly from 135 to 210 fps (frames per second) with changing the ternary weight threshold from 0 to 0.99.
We also performed qualitative analysis by simply drawing digits by hand and then observing the networks prediction, a sample of which are shown in figure \ref{fig:custom_digits}.


\section{ Car Tracking}

In addition to MNIST we tested CNN inference on a localization task detecting a toy car on a play mat from a bird's eye view. To make the inference more robust to illumination differences we trained the network on edge images such as those of figure \ref{fig:car_estimation_examples}.
The visual clutter of the play mat forces the network into learning feature kernels specifically identifying the car itself rather than just relying on finding the brightest area of the input image.

\subsection{Training}

\begin{figure}[t]
    \centering
   \includegraphics[width=0.43\linewidth]{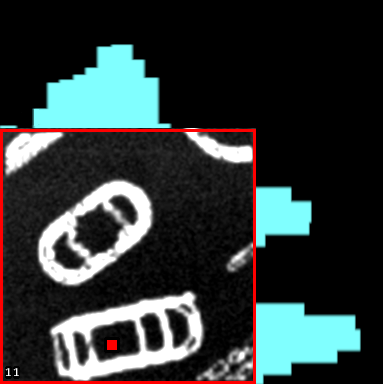}
   \includegraphics[width=0.43\linewidth]{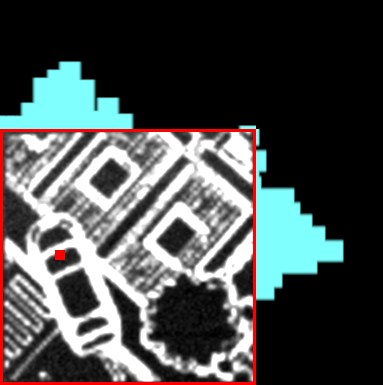}
     \caption{Two example outputs from the SCAMP system estimating a car's location within an image. Each input image is outlined in red, with the values of the neurons estimating x location plotted along the top and the neurons estimating y plotted along the right hand side. The estimated car location is indicated by the red dot within each image. \label{fig:car_estimation_examples}}
\end{figure}

Training images for this task were generated on the fly, two examples of these as seen through SCAMP5 are shown in figure \ref{fig:car_estimation_examples}.
Each training image was simply generated by drawing both the play mat and car at a random location and orientation, ensuring the car lay fully within the image bounds.
Additionally each training image was randomly scaled by $\pm10\%$ of it's original size, making the network more robust to small changes in scale.

Once again the trained network consisted of a single 5x5 kernel convolution layer of 16 filters, followed by a max-pooling layer into a fully connected output layer.
However this time the final layer consisted of 40 neurons, 20 representing potential x locations for the car (going from left to right) and the following 20 y locations (top to bottom).

Essentially the first 20 neurons divide the input image into 20 equally spaced vertical slices or "bins", with the activation value of each neuron representing the likely-hood that the car lies within it's associated bin.
We use equation \ref{eqn:binning activation} to then assign the correct output for these neurons and calculate the error used in back propagation, where $A^{x}_{n}$ is the correct activation of the $n^{th}$ neuron associated with x, and $i_{true_x}$ is the index of the neuron whose associated bin the car's true x location lies within.

\begin{equation}
    A^{x}_{n} = \frac{1}{1+\mid n - i_{true_x} \mid} 
    \label{eqn:binning activation}
\end{equation}

In much the same manner, the following 20 neurons are used to represent potential y locations for the car, each associated with a different horizontal bin, and whose correct output is assigned in much the same way as the neurons representing x locations.


\subsection{SCAMP5 Inference And Evaluation}

Inference evaluation for this task was again conducted by directly observing generated training images displayed on a monitor and then comparing the network's prediction against the car's true position for each image.
The index of the highest activated neuron for an x location was taken to represent the network's estimate for the car's x location $i_{est_x}$, with the estimate for y location $i_{est_y}$ defined in similar manner.
The indices of the neurons whose associated bins contain the true x and y location of the car are denoted by $i_{true_x},i_{true_y}$.
Error is then measured in terms of distance between the predicted and true locations as given by expression \ref{eqn:car err eqn}.

\begin{equation}
    \sqrt{(i_{est_x}-i_{true_x})^{2} + (i_{est_y}-i_{true_y})^{2}}
    \label{eqn:car err eqn}
\end{equation}

\begin{figure}[t]
    \centering
   \includegraphics[width=0.9\linewidth]{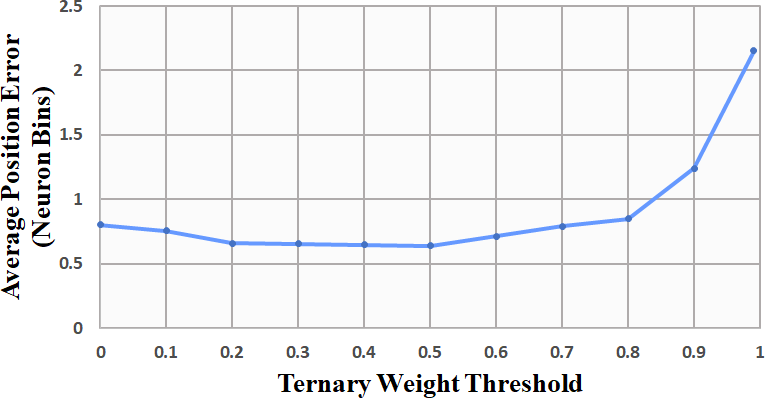}
   \caption{Plot showing estimation error against ternary weight threshold. \label{fig:car_results_graph}}
\end{figure}

Figure \ref{fig:car_results_graph} illustrates how the threshold used to determine the final ternary weights used during inference has a similar effect as it did upon the MNIST prediction accuracy in section \ref{subsec:MNIST results}.
Using a threshold of zero, resulting in binary weights, again actually results in a higher error than many other threshold values.
The error was seen to remain roughly the same between threshold values $0.2$ to $0.5$ after which an exponential increase in error is observed.
The computational cost varied in a same manner as \ref{subsec:MNIST results} as once again 5x5 kernels were used, however with the need to perform character extraction frame-rate was slightly increased, now ranging from 140 to 250 fps, for ternary weight thresholds 0 to 0.99.

\section{Conclusions and Future Directions \label{sec:conclusion and future}}
This presented an approach embedding CNNs directly into Pixel Processor Array sensors, which have the computational power and flexibility to make such developments possible. 
We focused on organizing computation to exploit the PPA's massively parallel processor array, computing convolutions in data-parallel fashion while optimizing the use of local memory resources. 
The results are promising. 
We were able to carry out the bulk of computation in the sensor, and successfully demonstrate good classification performance on example tasks. 
While our algorithms implement convolutions using digital logic operations, we have conducted our experiments using the current generation of PPA device prototypes (SCAMP-5), designed primarily for analog operation and fabricated using dated silicon technologies. 
While offering impressive performance and efficiency, the current technology used here is characterized by low clock speeds (10MHz) and moderate array size (256x256 pixels). We hope this work encourages development of next generation PPA devices, enabling the implementation of deeper networks. 
The development of architectures specifically designed to cater for visual processing is an important step in advancing and understanding Vision in general. We hope this work inspires others to consider the potential of this direction.

\section*{Data Access and Acknowledgements}
Supported by UK EPSRC EP/M019454/1 and EP/M019284/1. The nature of the PPA means that the data used for evaluation in this work is never recorded.

\newpage

{\small
\bibliographystyle{ieee_fullname}
\bibliography{main}
}

\end{document}